\documentclass[10pt,twocolumn,letterpaper]{article}

\usepackage{cvpr}              %

\usepackage{booktabs}
\usepackage{multirow}
\usepackage{float}

\definecolor{cvprblue}{rgb}{0.21,0.49,0.74}
\usepackage[pagebackref,breaklinks,colorlinks,allcolors=cvprblue]{hyperref}

\def\paperID{41} %
\def\confName{CV4Animals}
\def\confYear{2025}

\title{Domain-Adaptive Pretraining Improves Primate Behavior Recognition}

\author{Felix B. Mueller\textsuperscript{1} \qquad Timo Lueddecke\textsuperscript{1} \qquad Richard Vogg\textsuperscript{1} \qquad Alexander S. Ecker\textsuperscript{1,2}\\
\small\textsuperscript{1}Institute of Computer Science and Campus Institute Data Science, University of Göttingen, Germany\\
\small\textsuperscript{2}Max Planck Institute for Dynamics and Self-Organization, Göttingen, Germany\\
{\tt\small felix.mueller@cs.uni-goettingen.de}
}

\begin{document}
\maketitle
\begin{abstract}

Computer vision for animal behavior offers promising tools to aid research in ecology, cognition, and to support conservation efforts.   Video camera traps allow for large-scale data collection, but high labeling costs remain a bottleneck to creating large-scale datasets. We thus need data-efficient learning approaches.  In this work, we show that we can utilize self-supervised learning to considerably improve action recognition on primate behavior. On two datasets of great ape behavior (PanAf and ChimpACT), we outperform published state-of-the-art action recognition models by 6.1 \%pt.\ accuracy and 6.3 \%pt.\ mAP, respectively. We achieve this by utilizing a pretrained V-JEPA model and applying domain-adaptive pretraining (DAP), i.\,e.\ continuing the pretraining with in-domain data.  We show that most of the performance gain stems from the DAP. Our method promises great potential for improving the recognition of animal behavior, as DAP does not require labeled samples. Code is available at \url{https://github.com/ecker-lab/dap-behavior}
\end{abstract}

\section{Introduction}
\label{sec:intro}

There has been a growing interest in computer vision methods for animal behavior in recent years. Understanding animal behavior is crucial for many different fields, like cognition, ecology, and animal conservation. In this domain, non-human primate cognition is of high interest, due to its complexity and relation to human cognition.

\begin{figure}
    \centering
    \includegraphics[width=1\linewidth]{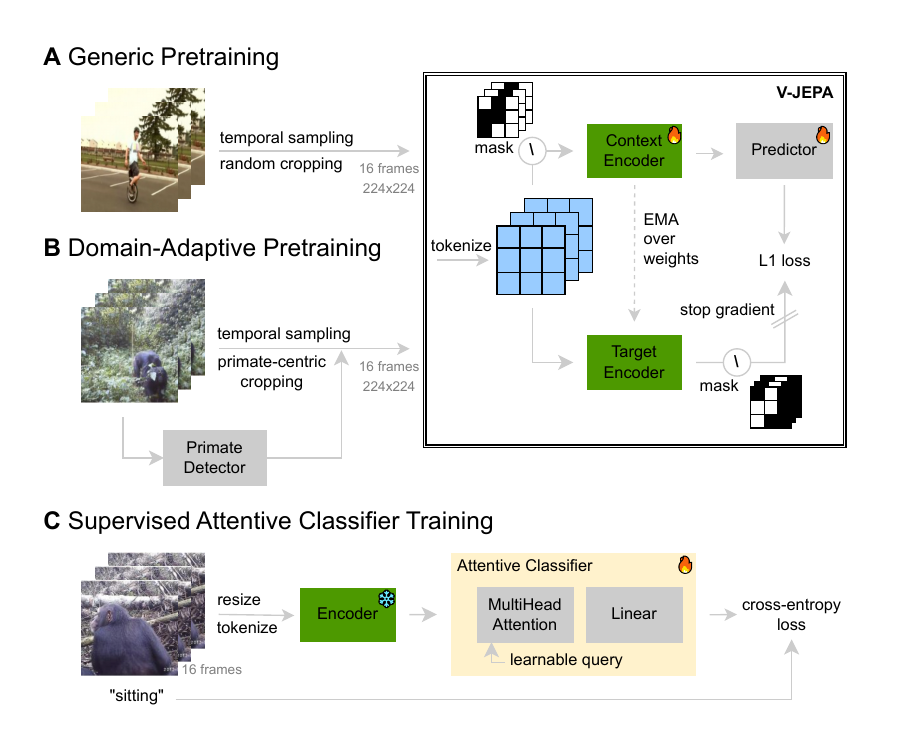}
    \caption{Overview over our methodology and the V-JEPA training paradigm. EMA: exponential moving average. \includegraphics[height=2ex]{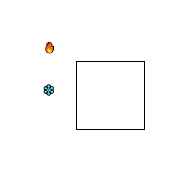}: trainable parameters. \includegraphics[height=2ex]{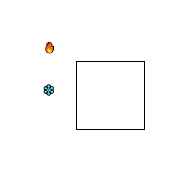}: frozen parameters. \textbf{A.} Self-supervised training utilizing human action datasets, done by \cite{bardes_revisiting_2024}. \textbf{B.} Our domain-adaptive pretraining utilizes the same learning paradigm as in A. \textbf{C.} We freeze the encoder from  A and B and train a classifier on top.} 
    \label{fig:overview}
\end{figure}

Animal behavior monitoring using computer vision is widely adopted in lab settings \cite{mathis_deeplabcut_2018,pereira2022sleap}, but these methods not easily transferable to more complex visual environments \cite{vogg_computer_2024}. They exploit the characteristics of one very constrained environment, which is often not possible in the wild. For in-the-wild settings, we are typically able to collect large amounts of data, but only parts of the data are relevant for the research problem at hand and a crucial limiting factor is labeling cost. This makes data-efficient machine learning methods highly relevant.

In this work, we explore self-supervised learning as a data-efficient learning method. In self-supervised learning, models are trained to learn useful representations of input data \textit{without requiring labeled data}. Recently, there has been considerable progress in video representation learning using masked autoencoding objectives \cite{bardes_revisiting_2024, tong_videomae_2022}. 

We apply self-supervised learning to the task of \textit{behavior recognition}, i.e. performing action recognition with primates as subjects. We perform domain-adaptive pretraining on a V-JEPA \cite{bardes_revisiting_2024} backbone already pretrained on human video data and then train an attentive classifier on top of the frozen backbone (Fig.~\ref{fig:overview}). 
We thereby beat the previous best published results on behavior recognition on PanAf500 \cite{brookes_panaf20k_2024} and ChimpACT \cite{ma_chimpact_2023}. On both datasets, we see that DAP gives large performance improvements compared to using the pretrained V-JEPA backbone directly.

\section{Related Work}
\label{sec:related_work}

Over the last years, a growing number of datasets for animal behavior recognition have been published, both general \cite{ng_animal_2022, rodriguez-juan_visual_2025} and primate specific. The latter range from zoo recordings \cite{ma_chimpact_2023, fuchs_forest_2024} over in-the-wild recordings using camera traps \cite{brookes_panaf20k_2024, brookes_panaf-fgbg_2025} to drone footage \cite{duporge_baboonland_2024}. 

There is a growing number of works on animal behavior recognition: Bain \etal \cite{bain_automated_2021} showed that they can discriminate two audiovisually distinctive actions in wild chimpanzees using deep learning. Various other works try to recognize primate behavior using deep learning \cite{fuchs_asbar_2023, lei_postural_2022, wiltshire_deepwild_2023}, improve behavior recognition with text descriptions~\cite{brookes_chimpvlm_2024}, or adapt the CLIP model~\cite{radford2021learning} for retrieval of camera trap events using natural language \cite{gabeff_wildclip_2024}. 

The goal in self-supervised learning is to 
learn useful representations from unlabeled data. This allows to perform a downstream task (e.g. ImageNet classification) without much additional training, either by training a classification head on top of the frozen backbone model or by finetuning for few iterations. While there exists a plethora of self-supervised learning methods for images \cite{chen_simple_2020, caron_emerging_2021, zhou_ibot_2022, oquab_dinov2_2024}, approaches for video input are more rare and are typically based on masked autoencoding. In masked autoencoding, the training objective is to predict masked regions of a video given the unmasked regions. The reconstruction loss can either be calculated directly in pixel space \cite{tong_videomae_2022} or in latent space as is done in V-JEPA \cite{bardes_revisiting_2024} and SIGMA \cite{salehi2024sigma}. Most recent self-supervised learning approaches use attention-based models, either Vision Transformer (ViT) \cite{dosovitskiy2020image} or a variation of it.

In this work, we focus on V-JEPA \cite{bardes_revisiting_2024} as video backbone, as it is based on a large, diverse pretraining dataset, utilizing HowTo100M, Kinetics, and Something-Something-v2.

\section{Methodology}
\label{sec:methodology}

\newcommand{\maskfn}{\text{Mask}}
\newcommand{\enct}{$\overline{E}$ }
\newcommand{\enc}{\overline{E}}
\newcommand{\softmax}{\text{softmax}}
\newcommand{\Real}{\mathbb{R}}

To perform primate behavior recognition, we take a pretrained V-JEPA backbone and perform domain-adaptive pretraining without labels to adjust it to the primate domain. We then train an attentive classifier on top of the frozen backbone model to perform action recognition. 

In this section, we describe the self-supervised pretraining setup (Section~\ref{subsec:ssl}), a primate-centric sampling strategy for the domain-adaptivce pretraining (Section~\ref{subsec:primate-centric-sampling}) and the attentive classifier used for action recognition (Section~\ref{subsec:behavior-recognition}).

\subsection{Self-Supervised Pretraining}
\label{subsec:ssl}

We use the V-JEPA model, which consists of a context encoder $E$, a target encoder $\overline{E}$, and a predictor $P$. 
V-JEPA works on a video tensor $x \in \Real^{16\times 224 \times 224}$ consisting of 16 frames sampled with a temporal stride of 4 and with a resolution of $224 \times 224$. The input is patchified into patches of size $2 \times 16 \times 16$, tokenized, and combined with sine-cosine positional embeddings to produce a set of $N=1568$ tokens $X \in \Real^{N \times d}$, with $d=1024$ being the token dimension.

During training, part of the input sequence is masked and the training objective is to predict the latent representation of masked tokens given the unmasked tokens (see Fig.~\ref{fig:overview}). Let $\maskfn$ and $\maskfn^C$ be functions that split a set of tokens into the set of masked and unmasked tokens, respectively. Then the V-JEPA pretraining loss for an unlabeled data sample $X$ is formulated as 
\begin{equation}
\label{eq:LMAE}
    L_{\text{MAE}} =\left\Vert P\left(E\left(\maskfn(X)\right)  \right) - \maskfn^C\left(\overline{E}(X)\right)\right\Vert_1.
\end{equation}
The parameters of context encoder $E$ and predictor $P$ (see Fig.~\ref{fig:overview}) are trained. To avoid collapse to trivial solutions, the parameters of target encoder $\overline{E}$ are not trained but computed as an exponential moving average over the past weights of the context encoder $E$. For generic pretraining, both teacher and student weights are initialized to the same random weights. For DAP, the final weights of teacher and student from generic pretraining are used.

The training objective $L_{\text{MAE}}$ was used during V-JEPA pretraining and is also used for the domain-adaptive pretraining we perform.

\subsection{Primate-Centric Sampling}
\label{subsec:primate-centric-sampling}

We find that primate datasets are less object-centric compared to common datasets for human action recognition, like Something-Something-v2. Videos might be almost empty with one primate at a distance or they might contain many primates at the same time. To capture fine-grained details of individual primates without increasing the input size (and therefore memory and runtime complexity) of Vision Transformer models, we preprocess our primate-specific datasets for domain-adaptive pretraining by using Grounding DINO \cite{liu_grounding_2024} as off-the-shelf open-vocabulary object detector to crop out parts of the video that contain primates (see Section~\ref{subsec:hyperparameters} for details).

We perform the domain-adaptive pretraining on three-second video snippets $v$ (see Section~\ref{subsec:dap-data} on how we chunk the in-domain pretraining data into video snippets). We crop, resize, and downsample $v$ both spatially and temporally to produce a video tensor $x$ for the self-supervised pretraining, 
\begin{equation}
    x = \text{Sample}_{16}(\text{CropResize}(v, b)).
\end{equation}
CropResize spatially crops all frames of the video snippet $v$ to a bounding box $b \in \Real^4$ and resizes them to $224\times 224$, while Sample\textsubscript{16} extracts a subsample of 16 frames. To crop spatially, we apply the primate detector on the center frame of $v$ and randomly select one of the predicted bounding boxes. We increase the bounding box edge lengths by a factor of 1.25 on each side and then crop with a random scale between 0.3 and 1.0 and a random aspect ratio between 0.75 and 1.35, yielding the bounding box $b$. We discard video snippets in which no primate was detected. To crop temporally, Sample$_{16}$ takes a random segment of 64 frames (2s) out of the three-second snippet, which it reduces to 16 frames by applying a temporal stride of 4.

\subsection{Supervised Attentive Classifier Training}
\label{subsec:behavior-recognition}

Our goal is to predict the behavior of an animal specified by a bounding box $b$ in an input video snippet $v$. We obtain a latent representation of the behavior by applying the pretrained encoder on a cropped video snippet
\begin{equation}
    h = \enc(\text{Sample}_{16}(\text{CropResize}(v, b)))
\end{equation}
with $h \in \Real^{N \times d}$.

We choose our attentive classifier to be a single multihead cross-attention layer which performs attention between a learnable query token $q \in \Real^{d}$ and the latent representations $h$ (keys and values), followed by a linear layer mapping to predicted class probabilities $\hat{y} \in [0, 1]^C$  with $C>1$. Given a latent representation $h$, we predict class probabilities $\hat{y}$ as 
\begin{equation}
\label{eq:AC}
    \hat{y} = \sigma(\text{Linear}(\text{CrossAttention}(q, h)))
\end{equation}
with $\sigma$ being the \textit{softmax} function for single-label classification and the \textit{sigmoid} function for multi-label classification. 
We use a cross entropy loss function for single-label ground truth $L_{1L} = -  \sum_{c=1}^C y_c \log \hat{y}$. In case of multi-label samples, we use binary cross entropy, $  L_{ML} = - \sum_{c=1}^C \left[ y_c \log \hat{y}_c + (1-y_c) \log (1-\hat{y}_c)\right]$,
with $y \in \{0, 1\}^C $ being the ground truth label for $(x, b)$. We optimize only the parameters of Linear and CrossAttention. \enct is frozen.

\vspace{-10pt}
\paragraph{Application to datasets.}

We use two datasets for action recognition: PanAf500 and ChimpACT. Both datasets provide ground truth  bounding box information as part of the evaluation protocol, but pose the behavior recognition task in two different ways: snippet-wise or frame-wise. 

For \textit{snippet-wise} classification (like PanAf500), the input is a video snippet $v$, a sequence of bounding $(b_1, \dots, b_{|v|})$ specifying how the animal moves through $v$, and one ground truth label $y$. To perform classification with our approach, we set $b$ to be the union over the bounding boxes $(b_1, \dots, b_{|v|})$. The PanAf evaluation video snippets are only 16 frames long, so we use no temporal stride for this dataset and directly input the spatially cropped snippets into the model.

For \textit{frame-wise} classification (like ChimpACT), the input is a video $v^*$, one point in time $t$, one bounding box $b$ at time $t$, and one ground truth label $y$ for this bounding box. To perform classification with our approach, we select a 2s video snippet $v$ from $v*$ centered at time $t$.

\section{Experiments}
\label{sec:experiments}

\subsection{Datasets and Evaluation Setup}
\label{subsec:eval-setup}
\label{subsec:dap-data}

We evaluate our approach on two video datasets of primates, PanAf \cite{brookes_panaf20k_2024} and ChimpACT \cite{ma_chimpact_2023}.

The \textbf{PanAf} dataset contains camera trap videos from 18 field sites in tropical Africa capturing chimpanzees and gorillas. The dataset consists of 20,000 coarsely annotated videos (PanAf20k) and 500 videos with fine-grained bounding box, track and frame-wise behavior annotation (PanAf500). Each video is 15\,s long. PanAf500 has frame-wise and per-individual single-label annotations of nine classes: sitting, walking, standing, hanging, climbing up, sitting on back, running, camera interaction, and climbing down (in descending order of frequency). %

Following the PanAf action recognition evaluation protocol, we train and evaluate the attentive classifier on 16-frame snippets of sustained single behavior of a single ape and predict a single behavior label for each snippet. Performance is evaluated using Top-1 Accuracy and average per-class accuracy.

The \textbf{ChimpACT} dataset contains 2\,h of video footage of zoo-housed chimpanzees recorded at the Leipzig Zoo. The recordings focus on one specific individual (Azibo) over the span of four years, but the dataset captures the whole chimpanzee group consisting of over 20 individuals. The dataset contains both indoor and outdoor recordings. ChimpACT has 23 behaviors annotated, which are be grouped into locomotion, social interaction, object interaction, and other. ChimpACT has ground truth multi-label behavior annotations for each ape in each frame. We evaluate mean average precision (mAP) following the AVA protocol \cite{gu_ava_2017}.

\vspace{-10pt}
\paragraph{Pretraining Data.}

For PanAf500, we utilize both PanAf500 and PanAf20k for pretraining, as PanAf20k is in-domain. For ChimpACT, we pretrain on all ChimpACT videos. In both cases we do not utilize any label information during pretraining even if available (all primate bounding boxes are obtained using the primate detector). We create samples by splitting all videos in the respective datasets into three-second snippets. We use a stride of 1.5s for PanAf500+20k, yielding 155k samples, and a stride of 1s for ChimpACT, yielding 6k samples.

\subsection{Architectural Choices}
\label{subsec:hyperparameters}

Following \cite{bardes_revisiting_2024}, we choose a ViT-L model with video input for $E$  and a narrow 12-layer transformer for the V-JEPA predictor $P$. $E$ has 304M parameters, $P$ has 22M parameters. All parameters are fine-tuned during domain-adaptive pretraining (DAP), which we perform for 14.4k steps with an effective batch size of 80, resulting in 1.2M samples seen during DAP. We choose a learning rate of $6\times 10^{-6}$ with warmup and cosine annealing. Following \cite{bardes_revisiting_2024}, we increase the weight decay during training from $0.01$ to $0.1$, instead of decreasing it.  DAP takes around 3 hours on 4 A100 GPUs.

We use zero-shot Grounding Dino \cite{liu_grounding_2024} for primate detection (82.7 mAP$_\text{All}$ on the PanAf500 primate detection task) as it can be applied easily to different datasets. We find the prompt “monkey.primate.ape.” and a box detection threshold of 0.2 to work well.

The attentive classifier has 4M parameters (Eq.~\ref{eq:AC}). When training it, we freeze the encoder $\enc$. We train for up to 30 epochs on the labeled train sets and perform early stopping based on validation accuracy. We report results on the test set, calculating class scores on a single view per test sample.

\section{Results}
\label{subsec:results}

\begin{table}
    \begin{tabular}{lrr}
    \toprule
    \textbf{Method} & \textbf{Top-1 Acc (\%)}& \textbf{C-Avg Acc (\%)}\\
    \midrule
    X3D \hfill \cite{DBLP:conf/cvpr/Feichtenhofer20}&  80.04&  56.10\\
    I3D \hfill \cite{DBLP:conf/cvpr/CarreiraZ17}&  79.29&  42.15\\
    3D ResNet-50 \hfill \cite{DBLP:conf/iccvw/HaraKS17}&  77.45&  55.17\\
    MViTV2 \hfill \cite{DBLP:conf/cvpr/LiW0MXMF22}&  81.09&  54.91\\
 TimeSformer \hfill \cite{DBLP:conf/icml/BertasiusWT21}& 79.45& \underline{56.38}\\
 
    \midrule
    V-JEPA (no DAP)& \underline{83.68}& \textbf{57.75}\\
    V-JEPA (DAP)&  \textbf{87.24}&  \underline{56.37}\\
    \bottomrule
    \end{tabular}
    \caption{Results on the behavioral action recognition task on the \textbf{PanAf500} test set. Baseline results are taken from \cite{brookes_panaf20k_2024}. For each baseline method we chose the best-performing setup based on Top-1 accuracy. DAP: Domain-adaptive pretraining. C-Avg Acc: Average per-class accuracy. We implemented the evaluation setup based on the description in \cite{brookes_panaf20k_2024}, see Section~\ref{subsec:eval-setup}.}
    \label{tab:results_panaf}
\end{table}

V-JEPA with domain-adaptive pretraining (DAP) outperforms plain V-JEPA by 3.5 percentage points in top-1 accuracy on PanAf500 (Table~\ref{tab:results_panaf}). Both V-JEPA with and without DAP outperform the previous best model. Even though we did not take specific measures to handle class imbalance, we still achieve competitive class-average accuracy. On the ChimpACT dataset we see the same pattern (Table~\ref{tab:results_chimpact}): domain-adaptive pretraining clearly outperforms standard V-JEPA on mAP. Both V-JEPA with and without DAP outperform the previous best models.

\begin{table}[t]
    \centering
    \begin{tabular}{lrrrr}
    \toprule
    \textbf{Method} & \textbf{mAP} & \textbf{mAP$_L$} & \textbf{mAP$_O$} & \textbf{mAP$_S$} \\
    \midrule
    ACRN \hfill \cite{DBLP:conf/eccv/SunSVMSS18}&  24.4&  \underline{58.7}&  \textbf{33.8}&  14.7\\
    LFB \hfill \cite{DBLP:conf/cvpr/WuF0HKG19}&  23.2&  45.0&  31.2&  {17.7}\\
    SlowOnly \hfill \cite{DBLP:conf/iccv/Feichtenhofer0M19}&  22.3&  52.3&  31.2&  13.8\\
    SlowFast \hfill \cite{DBLP:conf/iccv/Feichtenhofer0M19}&  24.3&  56.8&  31.5&  15.6\\
    {\color{gray}AlphaChimp$^\ast$} \hfill \cite{ma2024alphachimptrackingbehaviorrecognition} & {\color{gray}\textbf{34.3}} & {\color{gray}50.3} & {\color{gray}31.3} & {\color{gray}\textbf{29.3}} \\ 
    \midrule
    V-JEPA (no DAP)& {26.0} & \textbf{60.7} & 29.2 & {17.8}\\
    V-JEPA (DAP) & \underline{30.7} & 55.7 & \underline{33.4} & \underline{26.8}\\
    \bottomrule
    \end{tabular}
    \caption{Results on the spatiotemporal action detection track on the \textbf{ChimpACT} test set. Baseline results are taken from \cite{ma_chimpact_2023}, for each baseline method we chose the best-performing setup reported. DAP: Domain-adaptive pretraining. mAP: mean of average precision over classes. We report both the overall mAP and the mAP on \underline{L}ocomotion, \underline{O}bject interaction and \underline{S}ocial interaction. $^\ast$AlphaChimp is concurrent unpublished work.}
    \label{tab:results_chimpact}
\end{table}

\subsection{Ablations}
\label{subsec:ablations}

The V-JEPA attentive classifier is quite heavy,  using  multihead cross-attention followed by a two-layer MLP, and a final projection layer, resulting in 12M parameters~\cite{bardes_revisiting_2024}. We simplified it by dropping the MLP, which results in only a third of the  parameters required while producing similar or better performance (Table~\ref{tab:ablation_head_size}). We assume that the attentive classifier is overparametrized given the small evaluation dataset size and thus reducing the number of parameters reduces overfitting.

\begin{table}
    \centering
    \begin{tabular}{lrrrrr}
    \toprule
         &\multirow{3}{1cm}{\textbf{Param Count}}&  \multicolumn{2}{c}{\textbf{PanAf500}}&\multicolumn{2}{c}{\textbf{ChimpACT}}\\
         && \multicolumn{2}{c}{Top-1 Acc \%}&\multicolumn{2}{c}{mAP \%} \\
         \cmidrule{3-6}
         &&  no DAP&   DAP& no DAP&DAP\\
         \midrule
         Ours &4M &  \textbf{83.7}&   \textbf{87.2}&25.97&\textbf{30.74}\\
         V-JEPA &12M &  82.5&   87.1&\textbf{27.15}&29.74\\
         \bottomrule
    \end{tabular}
    \caption{We compare our simplified attentive classifier using only multihead attention to the V-JEPA classifier using a full transformer block.}
    \label{tab:ablation_head_size}
\end{table}

\section{Conclusion}
\label{sec:conclusion}

Our results show that domain-adaptive pretraining is a helpful tool when utilizing large pretrained backbone models for animal behavior tasks. A few GPU hours are enough to considerably improve downstream performance using DAP.

A limitation of our work is that we crop input videos to bounding boxes. While this allows to utilize  the limited input resolution of ViTs better, it disregards global context and necessitates one forward pass per region of interest. 

Self-supervised training allows us to utilize vast amounts of unlabeled data. Future work should explore to scale up DAP. Combining large and diverse sources of animal data might learn better representations and aid behavior recognition. This might especially improve performance on tail classes for which it might be hard to learn good representations from small datasets.

\section*{Acknowledgments}

The project was funded by the Deutsche Forschungsgemeinschaft (DFG, German Research Foundation) -- Project-ID 454648639 -- SFB 1528. The authors gratefully acknowledge the computing time granted by the Resource Allocation Board and provided on the supercomputer Emmy/Grete at NHR-Nord@Göttingen as part of the NHR infrastructure. The calculations for this research were conducted with computing resources under the project nib00021.

{
    \small
    \bibliographystyle{ieeenat_fullname}
    \bibliography{main}

\begin{thebibliography}{36}
\providecommand{\natexlab}[1]{#1}
\providecommand{\url}[1]{\texttt{#1}}
\expandafter\ifx\csname urlstyle\endcsname\relax
  \providecommand{\doi}[1]{doi: #1}\else
  \providecommand{\doi}{doi: \begingroup \urlstyle{rm}\Url}\fi

\bibitem[Bain et~al.()Bain, Nagrani, Schofield, Berdugo, Bessa, Owen, Hockings,
  Matsuzawa, Hayashi, Biro, Carvalho, and Zisserman]{bain_automated_2021}
Max Bain, Arsha Nagrani, Daniel Schofield, Sophie Berdugo, Joana Bessa, Jake
  Owen, Kimberley~J. Hockings, Tetsuro Matsuzawa, Misato Hayashi, Dora Biro,
  Susana Carvalho, and Andrew Zisserman.
\newblock Automated audiovisual behavior recognition in wild primates.
\newblock 7\penalty0 (46):\penalty0 eabi4883.
\newblock Publisher: American Association for the Advancement of Science.

\bibitem[Bardes et~al.()Bardes, Garrido, Ponce, Chen, Rabbat, {LeCun}, Assran,
  and Ballas]{bardes_revisiting_2024}
Adrien Bardes, Quentin Garrido, Jean Ponce, Xinlei Chen, Michael Rabbat, Yann
  {LeCun}, Mahmoud Assran, and Nicolas Ballas.
\newblock Revisiting feature prediction for learning visual representations
  from video.
\newblock V-{JEPA}.

\bibitem[Bertasius et~al.(2021)Bertasius, Wang, and
  Torresani]{DBLP:conf/icml/BertasiusWT21}
Gedas Bertasius, Heng Wang, and Lorenzo Torresani.
\newblock Is space-time attention all you need for video understanding?
\newblock In \emph{Proceedings of the 38th International Conference on Machine
  Learning, {ICML} 2021, 18-24 July 2021, Virtual Event}, pages 813--824.
  {PMLR}, 2021.

\bibitem[Brookes et~al.({\natexlab{a}})Brookes, Kukushkin, Mirmehdi, Stephens,
  Dieguez, Hicks, Jones, Lee, {McCarthy}, Meier, Normand, Wessling, Wittig,
  Langergraber, Zuberbühler, Boesch, Schmid, Arandjelovic, Kühl, and
  Burghardt]{brookes_panaf-fgbg_2025}
Otto Brookes, Maksim Kukushkin, Majid Mirmehdi, Colleen Stephens, Paula
  Dieguez, Thurston~C. Hicks, Sorrel Jones, Kevin Lee, Maureen~S. {McCarthy},
  Amelia Meier, Emmanuelle Normand, Erin~G. Wessling, Roman~M. Wittig, Kevin
  Langergraber, Klaus Zuberbühler, Lukas Boesch, Thomas Schmid, Mimi
  Arandjelovic, Hjalmar Kühl, and Tilo Burghardt.
\newblock The {PanAf}-{FGBG} dataset: Understanding the impact of backgrounds
  in wildlife behaviour recognition, {\natexlab{a}}.

\bibitem[Brookes et~al.({\natexlab{b}})Brookes, Mirmehdi, Kuhl, and
  Burghardt]{brookes_chimpvlm_2024}
Otto Brookes, Majid Mirmehdi, Hjalmar Kuhl, and Tilo Burghardt.
\newblock {ChimpVLM}: Ethogram-enhanced chimpanzee behaviour recognition,
  {\natexlab{b}}.

\bibitem[Brookes et~al.({\natexlab{c}})Brookes, Mirmehdi, Stephens, Angedakin,
  Corogenes, Dowd, Dieguez, Hicks, Jones, Lee, Leinert, Lapuente, {McCarthy},
  Meier, Murai, Normand, Vergnes, Wessling, Wittig, Langergraber, Maldonado,
  Yang, Zuberbühler, Boesch, Arandjelovic, Kühl, and
  Burghardt]{brookes_panaf20k_2024}
Otto Brookes, Majid Mirmehdi, Colleen Stephens, Samuel Angedakin, Katherine
  Corogenes, Dervla Dowd, Paula Dieguez, Thurston~C. Hicks, Sorrel Jones, Kevin
  Lee, Vera Leinert, Juan Lapuente, Maureen~S. {McCarthy}, Amelia Meier, Mizuki
  Murai, Emmanuelle Normand, Virginie Vergnes, Erin~G. Wessling, Roman~M.
  Wittig, Kevin Langergraber, Nuria Maldonado, Xinyu Yang, Klaus Zuberbühler,
  Christophe Boesch, Mimi Arandjelovic, Hjalmar Kühl, and Tilo Burghardt.
\newblock {PanAf}20k: A large video dataset for wild ape detection and
  behaviour recognition.
\newblock {\natexlab{c}}.

\bibitem[Caron et~al.()Caron, Touvron, Misra, Jégou, Mairal, Bojanowski, and
  Joulin]{caron_emerging_2021}
Mathilde Caron, Hugo Touvron, Ishan Misra, Hervé Jégou, Julien Mairal, Piotr
  Bojanowski, and Armand Joulin.
\newblock Emerging properties in self-supervised vision transformers.
\newblock {DINO}.

\bibitem[Carreira and Zisserman(2017)]{DBLP:conf/cvpr/CarreiraZ17}
Jo{\~{a}}o Carreira and Andrew Zisserman.
\newblock Quo vadis, action recognition? {A} new model and the kinetics
  dataset.
\newblock In \emph{2017 {IEEE} Conference on Computer Vision and Pattern
  Recognition, {CVPR} 2017, Honolulu, HI, USA, July 21-26, 2017}, pages
  4724--4733. {IEEE} Computer Society, 2017.

\bibitem[Chen et~al.()Chen, Kornblith, Norouzi, and Hinton]{chen_simple_2020}
Ting Chen, Simon Kornblith, Mohammad Norouzi, and Geoffrey Hinton.
\newblock A simple framework for contrastive learning of visual
  representations.
\newblock {SimCLR}, partially read.

\bibitem[Dosovitskiy et~al.(2020)Dosovitskiy, Beyer, Kolesnikov, Weissenborn,
  Zhai, Unterthiner, Dehghani, Minderer, Heigold, Gelly,
  et~al.]{dosovitskiy2020image}
Alexey Dosovitskiy, Lucas Beyer, Alexander Kolesnikov, Dirk Weissenborn,
  Xiaohua Zhai, Thomas Unterthiner, Mostafa Dehghani, Matthias Minderer, Georg
  Heigold, Sylvain Gelly, et~al.
\newblock An image is worth 16x16 words: Transformers for image recognition at
  scale.
\newblock \emph{arXiv preprint arXiv:2010.11929}, 2020.

\bibitem[Duporge et~al.()Duporge, Kholiavchenko, Harel, Wolf, Rubenstein,
  Crofoot, Berger-Wolf, Lee, Barreau, Kline, Ramirez, and
  Stewart]{duporge_baboonland_2024}
Isla Duporge, Maksim Kholiavchenko, Roi Harel, Scott Wolf, Dan Rubenstein, Meg
  Crofoot, Tanya Berger-Wolf, Stephen Lee, Julie Barreau, Jenna Kline, Michelle
  Ramirez, and Charles Stewart.
\newblock {BaboonLand} dataset: Tracking primates in the wild and automating
  behaviour recognition from drone videos.

\bibitem[Feichtenhofer(2020)]{DBLP:conf/cvpr/Feichtenhofer20}
Christoph Feichtenhofer.
\newblock {X3D:} expanding architectures for efficient video recognition.
\newblock In \emph{2020 {IEEE/CVF} Conference on Computer Vision and Pattern
  Recognition, {CVPR} 2020, Seattle, WA, USA, June 13-19, 2020}, pages
  200--210. Computer Vision Foundation / {IEEE}, 2020.

\bibitem[Feichtenhofer et~al.(2019)Feichtenhofer, Fan, Malik, and
  He]{DBLP:conf/iccv/Feichtenhofer0M19}
Christoph Feichtenhofer, Haoqi Fan, Jitendra Malik, and Kaiming He.
\newblock Slowfast networks for video recognition.
\newblock In \emph{2019 {IEEE/CVF} International Conference on Computer Vision,
  {ICCV} 2019, Seoul, Korea (South), October 27 - November 2, 2019}, pages
  6201--6210. {IEEE}, 2019.

\bibitem[Fuchs et~al.({\natexlab{a}})Fuchs, Genty, Bangerter, Zuberbühler, and
  Cotofrei]{fuchs_forest_2024}
Michael Fuchs, Emilie Genty, Adrian Bangerter, Klaus Zuberbühler, and Paul
  Cotofrei.
\newblock From forest to zoo: Great ape behavior recognition with
  {ChimpBehave}, {\natexlab{a}}.

\bibitem[Fuchs et~al.({\natexlab{b}})Fuchs, Genty, Zuberbühler, and
  Cotofrei]{fuchs_asbar_2023}
Michael Fuchs, Emilie Genty, Klaus Zuberbühler, and Paul Cotofrei.
\newblock {ASBAR}: an animal skeleton-based action recognition framework.
  recognizing great ape behaviors in the wild using pose estimation with domain
  adaptation, {\natexlab{b}}.
\newblock Pages: 2023.09.24.559236 Section: New Results.

\bibitem[Gabeff et~al.()Gabeff, Rußwurm, Tuia, and
  Mathis]{gabeff_wildclip_2024}
Valentin Gabeff, Marc Rußwurm, Devis Tuia, and Alexander Mathis.
\newblock {WildCLIP}: Scene and animal attribute retrieval from camera trap
  data with domain-adapted vision-language models.

\bibitem[Gu et~al.()Gu, Sun, Ross, Vondrick, Pantofaru, Li, Vijayanarasimhan,
  Toderici, Ricco, Sukthankar, Schmid, and Malik]{gu_ava_2017}
Chunhui Gu, Chen Sun, David~A. Ross, Carl Vondrick, Caroline Pantofaru, Yeqing
  Li, Sudheendra Vijayanarasimhan, George Toderici, Susanna Ricco, Rahul
  Sukthankar, Cordelia Schmid, and Jitendra Malik.
\newblock {AVA}: A video dataset of spatio-temporally localized atomic visual
  actions.

\bibitem[Hara et~al.(2017)Hara, Kataoka, and Satoh]{DBLP:conf/iccvw/HaraKS17}
Kensho Hara, Hirokatsu Kataoka, and Yutaka Satoh.
\newblock Learning spatio-temporal features with 3d residual networks for
  action recognition.
\newblock In \emph{2017 {IEEE} International Conference on Computer Vision
  Workshops, {ICCV} Workshops 2017, Venice, Italy, October 22-29, 2017}, pages
  3154--3160. {IEEE} Computer Society, 2017.

\bibitem[Lei et~al.()Lei, Dong, Guan, Xiang, Xie, Mu, Wang, and
  Ni]{lei_postural_2022}
Yujie Lei, Pengmei Dong, Yan Guan, Ying Xiang, Meng Xie, Jiong Mu, Yongzhao
  Wang, and Qingyong Ni.
\newblock Postural behavior recognition of captive nocturnal animals based on
  deep learning: a case study of bengal slow loris.
\newblock 12\penalty0 (1):\penalty0 7738.
\newblock Publisher: Nature Publishing Group.

\bibitem[Li et~al.(2022)Li, Wu, Fan, Mangalam, Xiong, Malik, and
  Feichtenhofer]{DBLP:conf/cvpr/LiW0MXMF22}
Yanghao Li, Chao{-}Yuan Wu, Haoqi Fan, Karttikeya Mangalam, Bo Xiong, Jitendra
  Malik, and Christoph Feichtenhofer.
\newblock Mvitv2: Improved multiscale vision transformers for classification
  and detection.
\newblock In \emph{{IEEE/CVF} Conference on Computer Vision and Pattern
  Recognition, {CVPR} 2022, New Orleans, LA, USA, June 18-24, 2022}, pages
  4794--4804. {IEEE}, 2022.

\bibitem[Liu et~al.()Liu, Zeng, Ren, Li, Zhang, Yang, Jiang, Li, Yang, Su, Zhu,
  and Zhang]{liu_grounding_2024}
Shilong Liu, Zhaoyang Zeng, Tianhe Ren, Feng Li, Hao Zhang, Jie Yang, Qing
  Jiang, Chunyuan Li, Jianwei Yang, Hang Su, Jun Zhu, and Lei Zhang.
\newblock Grounding {DINO}: Marrying {DINO} with grounded pre-training for
  open-set object detection.

\bibitem[Ma et~al.()Ma, Kaufhold, Su, Zhu, Terwilliger, Meza, Zhu, Rossano, and
  Wang]{ma_chimpact_2023}
Xiaoxuan Ma, Stephan~P. Kaufhold, Jiajun Su, Wentao Zhu, Jack Terwilliger,
  Andres Meza, Yixin Zhu, Federico Rossano, and Yizhou Wang.
\newblock {ChimpACT}: A longitudinal dataset for understanding chimpanzee
  behaviors.

\bibitem[Ma et~al.(2024)Ma, Lin, Xu, Kaufhold, Terwilliger, Meza, Zhu, Rossano,
  and Wang]{ma2024alphachimptrackingbehaviorrecognition}
Xiaoxuan Ma, Yutang Lin, Yuan Xu, Stephan~P. Kaufhold, Jack Terwilliger, Andres
  Meza, Yixin Zhu, Federico Rossano, and Yizhou Wang.
\newblock Alphachimp: Tracking and behavior recognition of chimpanzees, 2024.

\bibitem[Mathis et~al.()Mathis, Mamidanna, Cury, Abe, Murthy, Mathis, and
  Bethge]{mathis_deeplabcut_2018}
Alexander Mathis, Pranav Mamidanna, Kevin~M. Cury, Taiga Abe, Venkatesh~N.
  Murthy, Mackenzie~Weygandt Mathis, and Matthias Bethge.
\newblock {DeepLabCut}: markerless pose estimation of user-defined body parts
  with deep learning.
\newblock 21\penalty0 (9):\penalty0 1281--1289.
\newblock Publisher: Nature Publishing Group.

\bibitem[Ng et~al.()Ng, Ong, Zheng, Ni, Yeo, and Liu]{ng_animal_2022}
Xun~Long Ng, Kian~Eng Ong, Qichen Zheng, Yun Ni, Si~Yong Yeo, and Jun Liu.
\newblock Animal kingdom: A large and diverse dataset for animal behavior
  understanding.
\newblock pages 19023--19034.

\bibitem[Oquab et~al.()Oquab, Darcet, Moutakanni, Vo, Szafraniec, Khalidov,
  Fernandez, Haziza, Massa, El-Nouby, Assran, Ballas, Galuba, Howes, Huang, Li,
  Misra, Rabbat, Sharma, Synnaeve, Xu, Jegou, Mairal, Labatut, Joulin, and
  Bojanowski]{oquab_dinov2_2024}
Maxime Oquab, Timothée Darcet, Théo Moutakanni, Huy Vo, Marc Szafraniec,
  Vasil Khalidov, Pierre Fernandez, Daniel Haziza, Francisco Massa, Alaaeldin
  El-Nouby, Mahmoud Assran, Nicolas Ballas, Wojciech Galuba, Russell Howes,
  Po-Yao Huang, Shang-Wen Li, Ishan Misra, Michael Rabbat, Vasu Sharma, Gabriel
  Synnaeve, Hu Xu, Hervé Jegou, Julien Mairal, Patrick Labatut, Armand Joulin,
  and Piotr Bojanowski.
\newblock {DINOv}2: Learning robust visual features without supervision.

\bibitem[Pereira et~al.(2022)Pereira, Tabris, Matsliah, Turner, Li,
  Ravindranath, Papadoyannis, Normand, Deutsch, Wang, et~al.]{pereira2022sleap}
Talmo~D Pereira, Nathaniel Tabris, Arie Matsliah, David~M Turner, Junyu Li,
  Shruthi Ravindranath, Eleni~S Papadoyannis, Edna Normand, David~S Deutsch,
  Z~Yan Wang, et~al.
\newblock Sleap: A deep learning system for multi-animal pose tracking.
\newblock \emph{Nature methods}, 19\penalty0 (4):\penalty0 486--495, 2022.

\bibitem[Radford et~al.(2021)Radford, Kim, Hallacy, Ramesh, Goh, Agarwal,
  Sastry, Askell, Mishkin, Clark, et~al.]{radford2021learning}
Alec Radford, Jong~Wook Kim, Chris Hallacy, Aditya Ramesh, Gabriel Goh,
  Sandhini Agarwal, Girish Sastry, Amanda Askell, Pamela Mishkin, Jack Clark,
  et~al.
\newblock Learning transferable visual models from natural language
  supervision.
\newblock In \emph{International conference on machine learning}, pages
  8748--8763. PmLR, 2021.

\bibitem[Rodriguez-Juan et~al.()Rodriguez-Juan, Ortiz-Perez, Benavent-Lledo,
  Mulero-Pérez, Ruiz-Ponce, Orihuela-Torres, Garcia-Rodriguez, and
  Sebastián-González]{rodriguez-juan_visual_2025}
Javier Rodriguez-Juan, David Ortiz-Perez, Manuel Benavent-Lledo, David
  Mulero-Pérez, Pablo Ruiz-Ponce, Adrian Orihuela-Torres, Jose
  Garcia-Rodriguez, and Esther Sebastián-González.
\newblock Visual {WetlandBirds} dataset: Bird species identification and
  behavior recognition in videos.

\bibitem[Salehi et~al.(2024)Salehi, Dorkenwald, Thoker, Gavves, Snoek, and
  Asano]{salehi2024sigma}
Mohammadreza Salehi, Michael Dorkenwald, Fida~Mohammad Thoker, Efstratios
  Gavves, Cees~GM Snoek, and Yuki~M Asano.
\newblock Sigma: Sinkhorn-guided masked video modeling.
\newblock In \emph{European Conference on Computer Vision}, pages 293--312.
  Springer, 2024.

\bibitem[Sun et~al.(2018)Sun, Shrivastava, Vondrick, Murphy, Sukthankar, and
  Schmid]{DBLP:conf/eccv/SunSVMSS18}
Chen Sun, Abhinav Shrivastava, Carl Vondrick, Kevin Murphy, Rahul Sukthankar,
  and Cordelia Schmid.
\newblock Actor-centric relation network.
\newblock In \emph{Computer Vision - {ECCV} 2018 - 15th European Conference,
  Munich, Germany, September 8-14, 2018, Proceedings, Part {XI}}, pages
  335--351. Springer, 2018.

\bibitem[Tong et~al.()Tong, Song, Wang, and Wang]{tong_videomae_2022}
Zhan Tong, Yibing Song, Jue Wang, and Limin Wang.
\newblock {VideoMAE}: Masked autoencoders are data-efficient learners for
  self-supervised video pre-training.
\newblock 35:\penalty0 10078--10093.
\newblock {VideoMAE}.

\bibitem[Vogg et~al.()Vogg, Lüddecke, Henrich, Dey, Nuske, Hassler, Murphy,
  Fischer, Ostner, Schülke, Kappeler, Fichtel, Gail, Treue, Scherberger,
  Wörgötter, and Ecker]{vogg_computer_2024}
Richard Vogg, Timo Lüddecke, Jonathan Henrich, Sharmita Dey, Matthias Nuske,
  Valentin Hassler, Derek Murphy, Julia Fischer, Julia Ostner, Oliver Schülke,
  Peter~M. Kappeler, Claudia Fichtel, Alexander Gail, Stefan Treue, Hansjörg
  Scherberger, Florentin Wörgötter, and Alexander~S. Ecker.
\newblock Computer vision for primate behavior analysis in the wild.
\newblock version: 1.

\bibitem[Wiltshire et~al.()Wiltshire, Lewis-Cheetham, Komedová, Matsuzawa,
  Graham, and Hobaiter]{wiltshire_deepwild_2023}
Charlotte Wiltshire, James Lewis-Cheetham, Viola Komedová, Tetsuro Matsuzawa,
  Kirsty~E. Graham, and Catherine Hobaiter.
\newblock {DeepWild}: Application of the pose estimation tool {DeepLabCut} for
  behaviour tracking in wild chimpanzees and bonobos.
\newblock 92\penalty0 (8):\penalty0 1560--1574.
\newblock \_eprint:
  https://onlinelibrary.wiley.com/doi/pdf/10.1111/1365-2656.13932.

\bibitem[Wu et~al.(2019)Wu, Feichtenhofer, Fan, He, Kr{\"{a}}henb{\"{u}}hl, and
  Girshick]{DBLP:conf/cvpr/WuF0HKG19}
Chao{-}Yuan Wu, Christoph Feichtenhofer, Haoqi Fan, Kaiming He, Philipp
  Kr{\"{a}}henb{\"{u}}hl, and Ross~B. Girshick.
\newblock Long-term feature banks for detailed video understanding.
\newblock In \emph{{IEEE} Conference on Computer Vision and Pattern
  Recognition, {CVPR} 2019, Long Beach, CA, USA, June 16-20, 2019}, pages
  284--293. Computer Vision Foundation / {IEEE}, 2019.

\bibitem[Zhou et~al.()Zhou, Wei, Wang, Shen, Xie, Yuille, and
  Kong]{zhou_ibot_2022}
Jinghao Zhou, Chen Wei, Huiyu Wang, Wei Shen, Cihang Xie, Alan Yuille, and Tao
  Kong.
\newblock {iBOT}: Image {BERT} pre-training with online tokenizer.
\newblock {iBot}.

\end{thebibliography}
}

\end{document}